\documentclass{article}

\usepackage{microtype}
\usepackage{graphicx}
\usepackage{subfigure}
\usepackage{booktabs} 
\usepackage{amssymb}
\usepackage{amsmath}
\usepackage[utf8]{inputenc}
\usepackage[T1]{fontenc}

\usepackage{hyperref}

\usepackage[accepted]{icml2018t}

\icmltitlerunning{Accurate Kernel Learning for Linear Gaussian Markov processes}

\begin{document}


\twocolumn[
\icmltitle{Accurate Kernel Learning for Linear Gaussian Markov Processes\\
using a Scalable Likelihood Computation}




\begin{icmlauthorlist}
\icmlauthor{Stijn de Waele}{to}
\end{icmlauthorlist}

\icmlaffiliation{to}{ExxonMobil Research and Engineering Company, Annandale, New Jersey, USA}

\icmlcorrespondingauthor{Stijn de Waele}{stijn.dewaele@exxonmobil.com}

\icmlkeywords{Gaussian Processes, Kernel Learning, State-Space models, Autoregressive Models, Time Series Analysis}

\vskip 0.3in
]



\printAffiliationsAndNotice{}  

\begin{abstract}
We report an exact likelihood computation for Linear Gaussian Markov processes that
is more scalable than existing algorithms for complex models
and sparsely sampled signals. Better scaling is achieved through elimination of
repeated computations in the Kalman likelihood, and by using the diagonalized
form of the state transition equation. Using this efficient computation,
we study the accuracy of kernel learning using maximum likelihood and the posterior
mean in a simulation experiment. The posterior mean with a reference
prior is more accurate for complex models and sparse sampling.
Because of its lower computation load, the maximum likelihood estimator is
an attractive option for more densely sampled signals and lower order models.
We confirm estimator behavior in experimental data through
their application to speleothem data.
\end{abstract}

\section{Introduction}

Gaussian processes (GPs) are used in a wide range of applications
in science and engineering \cite{rasmussen2006gaussian}. A GP model
with a fixed kernel suffices for some applications. However, typically
it is required to learn the kernel from available data. Furthermore,
there is a trend towards more complex kernel parametrizations beyond
the basic models with a few parameter, such as the Matérn or squared
exponential kernels. An example of this trend is deep kernel learning,
where a neural network is used as part of the model \cite{wilson2016deep}.
This trend agrees well with results from the domain of time series
analysis, where it was found that simple models are adequate in describing
some processes, but models of moderate to high complexity are typically
preferred, and often critical to solve the problem at hand \cite{prado2010time}.

In many applications of Gaussian processes, kernel learning is performed
from irregularly spaced samples, either by experimental design, e.g.
in Bayesian optimization \cite{ghahramani2015probabilistic}, or as
a result of the sampling process, e.g. in climate data \cite{sinha2015trends}
and exoplanet detection in astronomy \cite{khan2017discovering}.
While irregular sampling offers the benefit of spectral estimation
beyond the average sampling frequency \cite{Broersen2007}, this type
of sampling also poses challenges for the statistical accuracy of
estimated kernels \cite{Broersen::2010}.

Motivated by these trends and challenges, we investigate the quality
of kernel learning algorithms from irregularly sampled data. Linear
Gaussian Markov models are an excellent practical choice for parametrization
of Gaussian processes with a one-dimensional index set (or: time series),
because of their wide applicability, computational convenience, and
availability of estimation algorithms \cite{prado2010time,murphy2012machine,rasmussen2006gaussian}.
As motivated in section \ref{sub:Linear-Markov-Models}, we focus
on two Markov models: the Linear-Gaussian State-Space model
and the Autoregressive model.

We compare the frequentist Maximum Likelihood (ML) estimate to the
Bayesian posterior mean. We do not use strong priors so that both
the ML and Bayesian estimate are only based on the model structure
and the data. In this way, the Bayesian estimate can be used as a
direct replacement for ML where desired. However, this work is also
relevant for the situation where we use stronger priors, because
in this situation we still have to decide between using the more
efficient posterior mode estimate versus other point estimates such
as the posterior mean. This choice is analogous to the choice between
the ML and posterior mean as discussed in this paper.

The benefit of the ML estimator is that it is considerably more computationally
efficient than Bayesian estimates. Many theoretical results exist
that show exact equivalence between ML and Bayesian point estimates
with uninformative priors for certain model structures, as well as
the general result that the estimates converge in the limit of large
samples \cite{bayarri2004interplay}. Furthermore, the ML estimate,
unlike the posterior mode, is independent of the chosen model parametrization,
a desirable property that it shares with preferred Bayesian point
estimates such as the posterior mean. However, \cite{Broersen::2010} reports that for the kernel learning problem the ML estimator can perform
poorly under certain conditions, resulting in very inaccurate models,
which manifests in the kernel spectrum as spurious peaks.

In an extensive simulation study, we quantify this problem under various
estimation conditions, and show that the posterior mean estimate is
successful in reducing the spurious peaks, resulting in more accurate
models. The simulation study is a key contribution of this paper,
because only a simulation study allows a meaningful comparison between the estimators.
The existing asymptotic theory does not describe the significant difference in performance
between the estimators that is observed in practice.

Computational complexity is a significant challenge for kernel learning
and is therefore an active area of research. For example, \cite{dong2017scalable}
reports an algorithm for an approximate likelihood computation that
scales linearly with the number of observations. As reported in a
early paper by Jones \cite{jones1980maximum}, the \textit{exact}
likelihood for Linear Gaussian State-Space models can be computed
with the same linear scaling; this result has been used for GP kernel
learning using Maximum Likelihood \cite{sarkka2013spatiotemporal,gilboa2015scaling}. In many
parameter estimation problems, usage of the exact likelihood as opposed
to an approximation leads to more accurate and robust estimators, and is
therefore preferable. For example, for the autoregressive model, the
usage of the approximate likelihood can result in non-stationary models,
whereas the exact ML estimator is guaranteed to result in stationary
models \cite{broersen2006automatic}. We introduce an improvement
of the exact likelihood computation that further reduces computational
cost and improves scalability with model complexity.

While we focus on parameter estimation from irregularly sampled data,
the presented algorithms can directly be used for the problem of missing
data. The reason is that the problem of irregular sampling is addressed
by rounding sampling times to a fine regular grid, effectively converting
the problem into a regularly sampled signal with a large fraction
of missing data.

Section \ref{sec:Definitions} provides definitions of the considered
models and error metric. Section \ref{sec:Scalable-exact-likelihood}
introduces the likelihood computation with improved scalability.
\ref{sec:Estimators} describes the estimators based on this likelihood computation.
Crucial to the quality evaluation of the estimators, section 
\ref{sec:Simulation-Experiments} describes the simulation experiments
comparing the proposed estimators.
Finally, the algorithms are applied to experimental speleothem data in section \ref{sec:monsoon}.


\section{\label{sec:Definitions}Definitions}

\subsection{Process}

We consider a zero mean, stationary Gaussian Markov process $y_{n}\in\mathbb{R}^{m}$
over the discrete one-dimensional index (time) variable $n\in\mathbb{Z}$.
The available observations of this process are $N_{a}$ irregularly
spaced observations $y_{n_{i}}$ at index $n_{i}$, taken over a measurement
interval of length $N$. The set of available index values is denoted
$\mathfrak{N}$. If a mean value or trend is present in a dataset,
it can be subtracted as preprocessing, and be added back to the predictions
made with the estimated model.

\subsection{\label{sub:Linear-Markov-Models}Linear Markov Models}

We define two types of Linear Markov models: the linear Gaussian
State-Space (LG-SS) model and the autoregressive (AR) model. Please
refer to \cite{prado2010time} and \cite{broersen2006automatic} for
some of the basic properties and results for these models that are
used in the remainder of this section.

Following the notation in \cite{murphy2012machine}, we write the
linear Gaussian State-Space model as:
\[
\begin{array}{lll}
z_{n} & = & Az_{n-1}+\epsilon_{n}\\
y_{n} & = & Cz_{n}+\delta_{n}
\end{array}
\]
with state vector $z_{n}\in\mathbb{R}^{s}$, matrices $A\in\mathbb{R}^{s\times s}$
and $C\in\mathbb{R}^{m\times s}$; $\epsilon_{n}\in\mathbb{R}^{s}$
and $\delta_{n}\in\mathbb{R}^{m}$ are normally distributed, temporally
uncorrelated stationary stochastic processes with covariance matrix
$Q\in\mathbb{R}^{s \times s}$, and $R\in\mathbb{R}^{m\times m}$,
respectively. The LG-SS model is the most general of the two models;
the autoregressive model can be rewritten as an equivalent LG-SS
model. Furthermore, this model is used to compute the Kalman
likelihood in section \ref{sec:Scalable-exact-likelihood}.

The autoregressive model of order $p$ is defined by:
\[
y_{n}=\sum_{i=1}^{p}a_{i}y_{n-1}+v_{n}
\]
where the $a_{i}$ are referred to as prediction coefficients, and
$v_{n}$is a normally distributed, temporally uncorrelated stationary
stochastic process with covariance matrix $V\in R^{m\times m}$, written
as $\sigma_{v}^{2}$ for $m=1$. 

We will now discuss some of the many practical and theoretical motivations
for the AR model. (i) Under mild conditions, a stationary Gaussian
process can be approximated arbitrary well by an AR($p$) model of
sufficiently high order, i.e. every process satisfying these conditions
has an AR($\infty$) representation. (ii) AR models of low to moderate
order $p$ are successfully used to model a wide range of processes:
the AR(1) model is the discrete-time version of the Matérn-3/2 kernel
or Ornstein\textendash Uhlenbeck process used in machine learning,
physics and economics; models of moderate order (5-30) are
successful in an even wider range of applications, see, e.g., \cite{zhang2017classification,ramadona2016prediction}.
Finally, (iii) AR models are used as the basis for a range of successful
reduced statistics Moving Average (MA) and Autoregressive-Moving average (ARMA) estimators.

Conversely, Maximum Likelihood estimation for the MA
and ARMA models is inaccurate
even for regularly spaced observations. For ARMA models, the theoretical
result explaining these problems is that the Cramèr-Rao lower bound
for the estimation error for even the simplest ARMA(1,1) model is
infinite. Because of the correspondence between ARMA models and LG-SS,
the same issue exist for general LG-SS parameter estimation \cite{auger2016state}.

Since the AR model parameter fully characterize the Gaussian process,
both the covariance function $k$, and the power spectral density
$h$ can be computed from the AR model parameters. An alternative
parametrization of the AR($p$) model are the partial autocorrelations
$\phi_{i}$. Because the requirement of stationarity can be simply
stated as $|\phi_{i}|<1$, this representation is central
in AR parameter estimation. Partial autocorrelations are also defined
for the vector AR process \cite{marple1987digital}.

\subsection{\label{sub:Error-metric}Error metric}

In general, models should not be evaluated on the difference between
estimated and true model parameters, because the impact of a given
difference in parameter values can have a vastly different impact
on model performance depending on the location in parameter space
where this difference is observed. Furthermore, it precludes comparing
models of a different structure. Rather, we should evaluate models
by evaluating their accuracy when used for inference. We evaluate
models using the model error ME \cite{broersen2006automatic}. We
will now summarize some key results for ME as needed for this paper.

The model error ME is a normalized version of the one-step-ahead prediction
error PE:
\[
\mbox{ME}=N_{a} \left( \frac{\mbox{PE}}{\sigma_{v}^{2}}-1 \right),
\]
where PE is the expectation of the one-step-ahead prediction error
of the estimated model compared to the generating process. Besides
this direct time domain interpretation as normalized one-step ahead
prediction error, it is asymptotically equivalent to the Spectral
Distortion. This equivalence motivates
reporting estimated models in the frequency domain using the log power spectral density.
Furthermore, the model error is asymptotically equivalent to the Kullback-Leibler
Divergence (KLD) for regularly sampled data. For regularly sampled
data, the asymptotic expectation of ME for estimated models is equal
to the number of estimated parameters: $\mbox{E}[\mbox{ME}]=p$.

The ME can be be computed efficiently for general ARMA (and consequently
for LG-SS) models. Because of its use in our simulation study, here we report
the expression for an estimated AR($p$) model $\hat{a}=(\hat{a}_{1},\hat{a}_{2},...,\hat{a}_{p})^{T}$
with respect to an AR($p)$ process with parameter vector $a$: 
\begin{equation}
\mbox{ME}(\hat{a},a)=N_{a}(\hat{a}-a)^{T}R^{-1}(\hat{a}-a),\label{eq:model-error-ar}
\end{equation}
where $R\in$$\mathbb{R}^{p\times p}$ is the covariance matrix
of $p$ consecutive observations of the true process $a$.

\section{\label{sec:Scalable-exact-likelihood}Scalable exact likelihood computation}

In this section we develop a scalable exact computation of the
likelihood for the LG-SS model that is more computationally efficient than the existing Kalman
likelihood. This  computation is used in the estimation algorithms described in section \ref{sec:Estimators}.

\subsection{Existing methods}

The likelihood for observations $y\in \mathbb{R}^{N_{a}}$, $y=(\begin{array}{cccc}
y_{n_{1}} & y_{n_{2}} & \cdots & y_{n_{Na}}\end{array})^{T}$ of a
Gaussian Process can be written as \cite{murphy2012machine}:
\[
L(y)=\mbox{\ensuremath{\mathcal{\log N}(0,K)}=}-\frac{1}{2}\log\left|K\right|-\frac{1}{2}y^{T}K^{-1}y+\mbox{constant}
\]
where $\mathcal{N}(\mu,S)$ is the multivariate Gaussian distribution,
and $K\in\mathbb{R}^{N_{a}\times N_{a}}$ is the data covariance matrix:
$K_{ij}=k(n_{i}-n_{j})$. We refer to this computation of the likelihood
for general covariance matrices $K$ as the covariance matrix (COVM)
method. This method is computationally expensive, i.e. $\mathcal{O}(N_{a}^{3})$,
and consequently its application is limited to small datasets.

The Kalman likelihood (KAL) is a more efficient, exact likelihood
computation for LG-SS models, which uses the decomposition of 
the likelihood into a sum of conditional likelihoods \cite{murphy2012machine}:
\begin{equation}
L(y)=\sum_{n\in\mathfrak{N}}\log\mathcal{N}(y_{n};\mu_{n},\Sigma_{n})
\end{equation}
where the mean vector $\mu_{n}\in\mathbb{R}^{s}$ and covariance matrix
$\Sigma_{n}\in\mathbb{R}^{s\times s}$ are computed using the Kalman
measurement equations for each $n\in\mathfrak{N}$, while the Kalman prediction
step is performed for all $N$ grid points. The Kalman likelihood for the initial state
distribution is given by the Lyapunov equation. This computation is $\mathcal{O}(N)$. Since it is often beneficial
to use a small grid time $T_{g}$, it is typical that $N_a\ll N$. Hence, we proceed to improve algorithm efficiency to achieve
an exact computation of $\mathcal{O}(N_{a})$.

\subsection{Kalman likelihood with precomputation}

For sparsely sampled data, it will occur frequently that no data is available for several
consecutive sample points. For the likelihood computation, this results in repeated application of the Kalman prediction step. In this section, we describe an algorithm
that precomputes elements of the repeated prediction step, thus reducing the overall computational complexity of the likelihood computation.

Performing $k$ prediction steps in absence of measurements yields:
\begin{equation}
\begin{array}{lll}
\mu_{n+k|n} & = & A^{k}\mu_{n}\\
\Sigma_{n+k|n} & = & F(A,\Sigma_{n},k)+G(A,Q,k)
\end{array}\label{eq:repeat-predict}
\end{equation}
where $\Sigma_n$ is the state covariance matrix conditional on all available preceding observations, and
\begin{align}
F(A,\Sigma_{n},k) & =A^{k}\Sigma_{n}(A^k)^T\nonumber\\
G(A,Q,k) & =\sum_{i=0}^{k-1}A^{i}Q(A^i)^{T}\nonumber
\end{align}
Computation of each of these contributions can be accelerated using precomputation
of components that only depend on the model parameters, i.e. $A^k$ and $G(A,Q,k)$.
We write $G$ in terms of the contributions to the sum:
$G=\sum_{i=0}^{k-1}g_{i}$,
with $g_{i}=A^{i}QA^{iT}$. For efficient computation of $g_{i}$
we use the recursive relations $g_{i}=Ag_{i-1}A^{T}$ and $G(A,Q,k)=G(A,Q,k-1)+g_{k-1}$.

After completion of the precomputation, we perform $N_a$ repeated prediction steps and measurement
steps, each of which has constant complexity for constant $s$. Therefore,
the computational complexity has been reduced to $\mathcal{O}(N_{a})$.
For AR models, the computation can be further accelerated
by exploiting the sparsity in $C$ and $Q$.

\subsection{Diagonalized Kalman likelihood with precomputation}

For more efficient computation of the matrix powers in the Kalman
likelihood for larger state dimension $s$, we decompose the
state vector $z$ according to the eigenbasis of $A$, yielding the
following equations for the state-space model:
\begin{eqnarray}
z_{n}^{e} & = & \Lambda z_{n-1}^{e}+\epsilon_{n}^{e}\nonumber \\
y_{n} & = & C^{e}z_{n}^{e}+\delta_{n}\nonumber
\end{eqnarray}
where $z^{e},\epsilon^{e}$ and $C^{e}$ are the state, process noise
and $C$ matrix, expressed on eigenbasis of $A$.

This decomposition requires that $A$ is diagonalizable. The $A$ matrix corresponding to an AR($p$) model is diagonalizable as long as $a_p\neq0$. If $a_p=0$, we
can remove trailing zero-valued coefficients to obtain an AR($p^*$) model with $p^*<p$, and $a_p^*\neq0$ which can be diagonalized. This model order reduction order does not affect the covariance function of $y$ and will therefore yield the correct likelihood value.

The measurement step is the same as in the original Kalman likelihood (KAL). The
repeated prediction step (eq. \ref{eq:repeat-predict}) becomes:
\[
\begin{array}{lll}
\mu_{n+k|n}^{e} & = & \Lambda^{k}\mu_{n}^{e}\\
\Sigma_{n+k|n}^{e} & = & \Lambda^{k}\Sigma_{n}\Lambda^{kH}+\sigma_{\epsilon}^{2}\sum_{i=0}^{k-1}\Lambda^{i}Q^{e}\Lambda^{iH}
\end{array}
\]
which is more computationally efficient to evaluate, because $\Lambda$
is diagonal. For the functions $F$ and $G,$ we find:
\[
F(\Lambda,\Sigma_{n},k)=F_{p}(\Lambda,k)\circ\Sigma_{n}
\]
where $\circ$ is the elementwise matrix product or Hadamard product, and $F_{p}(\Lambda,k)[i,j]=(\lambda_{i}\lambda_{j}^{*})^{s}$.
Since $F_{p}$ is data-independent, it can be precomputed. For the
second contribution $G$, we find:
\begin{equation}
G(\Lambda,Q^{e},k)=G_{p}(\Lambda,k)\circ Q^{e}\label{eq:sigma-pred-contrib-2}
\end{equation}
with
\[
G(\Lambda,k)[i,j]=\frac{1-(\lambda_{i}\lambda_{j}^{*})^{k}}{1-\lambda_{i}\lambda_{j}^{*}}
\]

Equation \ref{eq:sigma-pred-contrib-2} has reduced computational
complexity compared to the original expression because (i) it uses
the elementwise product, which is $\mathcal{O}(s^{2})$,
whereas the matrix multiplication is $\mathcal{O}(s^{3})$
and (ii) the usage of the geometric series allows for efficient exponentiation,
e.g. exponentiation by squaring.

Finally, the convariance of the unconditional distribution can also be computed more efficiently, because the Lyapunov equation reduces to a
per-element expression:
\[
\Sigma_{1|0}[i,j]=\frac{Q^{e}[i,j]}{1-\lambda_{i}\lambda_{j}^{*}}.
\]

\subsection{Computational load}

\begin{figure}[tbh]
\includegraphics{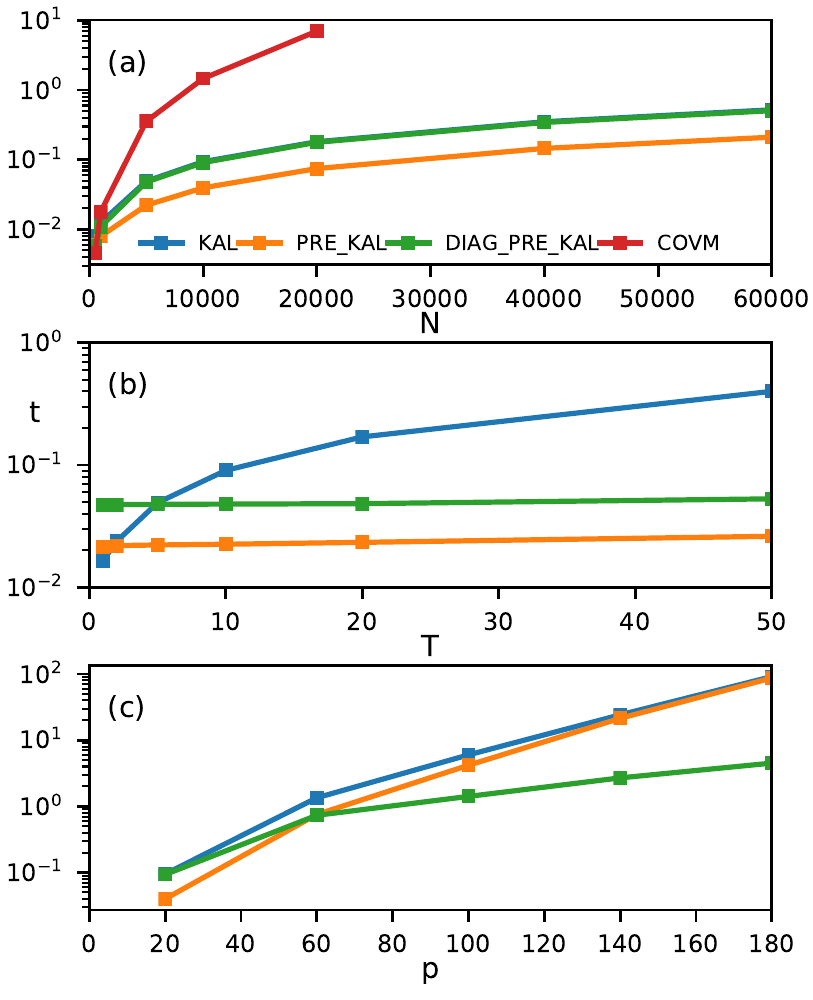}\protect\caption{\label{fig:Computational-load} Execution time in seconds for likelihood computation methods as a function of the signal duration (a); average
sampling interval (b) and model complexity (c).}
\end{figure}

The computational load of the likelihood computation is measured experimentally
on speech sample data from \cite{wilson2015kernel}. Irregularly sampled
data is derived by random subsampling. In figure \ref{fig:Computational-load}, we report the computational
load of the described likelihood computations implemented in Julia 0.6.0, and executed on a 2.60 GHz Intel Xeon E5-2670 CPU.

The experiments confirm the expected scaling behavior: Kalman-based
computations scale much better with signal duration $N$ (fig.\ref{fig:Computational-load}a);
precomputation methods PRE-KAL and DIAG-PRE-KAL are more efficient
for larger sampling intervals compared to the original Kalman likelihood
KAL (fig. \ref{fig:Computational-load}b). Finally, the usage of the
diagonalized state-space formulation (DIAG-PRE-KAL) scales better
with increasing model complexity (fig. \ref{fig:Computational-load}c).
However, because PRE-KAL uses only real-valued variables, it has a
lower memory footprint, which makes it more efficient for low-complexity
models.

\section{\label{sec:Estimators}Estimators}

In this section we describe our implementation of the maximum likelihood
and Bayesian point estimators for kernel learning for the autoregressive
(AR) model. We use the PRE-KAL algorithm, which is the most efficient for 
the considered model complexity, implemented in the probabilistic programming language Stan \cite{carpenter2016stan}. The Stan script is provided as supplementary material with this paper.

\subsection{\label{sub:round-to-grid}Setting grid and analysis time scales}

In case observations are made on a continuous index variable $t$,
the time indices are converted to integer indices $n_{i}$ by rounding the
continuous index variable to values on a regularly spaced grid with
spacing $T_{g}$. A coarser grid will result into larger errors in
the estimated process parameters. Considerations in setting this value
include (i) the time scale of interest (ii) process dynamics and (iii)
computational load and numerical stability.

An effective way to achieve accurate models at the time scale of interest
$T_{A}$ is to simply set the grid spacing $T_{g}$ equal to $T_{A}$.
However, this does not fully exploit all available data, and may introduce
errors in the process dynamics that are too large. In this case, we
can use modeling at interval $T_{A}$ \cite{dewaele2000order}. The
discrete-time signal can be split in $T_{A}/T_{g}$ segments of data
that can be treated independently, which means that the likelihood
can be computed as the sum of the likelihood per segment.

\subsection{\label{sub:Maximum-Likelihood-Estimator}Maximum Likelihood (ML)
Estimator}

Maximization of the likelihood is performed over the partial autocorrelations
$\phi$ using the L-BFGS algorithm, initiated from a number of different
starting points to reduce the probability of finding a local optimum.
Three starting points are obtained using the Burg AR parameter estimator
applied to regularly sampled signals derived from the original data
(i) using Nearest Neighbor interpolation (ii) using Linear Interpolation
and (iii) by ignoring missing data. Finally, (iv) the white noise model, i.e. $\phi_{i}=0$ for $i\in[1,p]$, is used as a starting point. The starting point resulting in the highest likelihood is used as the final estimate.

\subsection{Bayesian point estimate with uninformative prior (PMEAN)}

In addition to the ML estimate, we propose a Bayesian point estimate
for the model parameters. While it is possible and perfectly valid
to use the generated parameter samples for further inference,
we use a spoint estimate here because of the following reasons: (i) In
practice a posterior is rarely the end result of the analysis.
Rather, the samples are used to draw a conclusion that can be
formulated as a Bayesian decision problem. A point estimate can also
be interpreted as a result of a decision problem \cite{gelman2014bayesian};
(ii) the point estimate allows a direct comparison with the ML estimator,
and can be used as a direct replacement for it. Finally (iii) usage of a point estimate greatly reduces the computational complexity of subsequent application of the estimated kernel parameters, e.g. prediction, optimization or interpolation.

We approximate an uninformative prior by application of the AR(1)
reference prior \cite{berger1994noninformative} to the partial autocorrelations
$\phi_{i}$: $p(\phi_{i})\propto(1-\phi_{i}^{2})^{-1/2}$. The approximate location parametrization is given by $\kappa_{i}=\arcsin\phi_{i},\;-\pi<\kappa_{i}<\pi$. By definition, this is the parametrization where the reference prior is uniform \cite{bernardo2005reference}.

For many estimation problems, asymptotical theory accurately describes
estimator performance. In this regime, the posterior is narrow, and
various point estimators converge to the same estimate, including the posterior
mode, posterior mean, and the ML estimate \cite{gelman2014bayesian}.
Notably, because in the asymptotic regime $\mbox{E}[f(\theta)]\approx f(\mbox{E}[\theta])$,
the posterior mean of different parameterizations are equivalent.
However, for irregularly sampled data, we are typically not in this
regime and therefore have to be more precise in defining a Bayesian
point estimate. Because the model error ME (eq. \ref{eq:model-error-ar})
is approximately quadratic in the prediction parameters $\hat{a}$,
we use the posterior mean of $\hat{a}$ as point estimate: $\hat{a}=\mbox{E}[a|y]$,
which is the optimal Bayesian decision for a quadratic loss in $\hat{a}$.
This estimator is referred to as PMEAN. Posterior samples are drawn
with the Stan implementation of Hamiltonian Monte Carlo (HMC), initiated from the ML estimate.

\section{\label{sec:Simulation-Experiments}Simulation study}

Simulation experiments are critical for performance evaluation of kernel learning algorithms,
because the asymptotic results that can be obtained theoretically
are quite different from estimator performance in practice. In section \ref{sub:Case-study}, we quantify the impact
on the model error of the occurrence of spurious spectral peaks. In section
\ref{sub:Performance-range}, we determine the dependence of the model
error on various process and model parameters.

\subsection{\label{sub:Case-study}Case study various kernels}

The case study kernels are defined as follows:
\begin{itemize}
\item \textbf{case A} : exponential kernel with covariance function $\rho(r)=\exp(-r/200)$
, or, equivalently, an AR(1) process with parameter $\phi_{1}=a_{1}=\exp(-1/200)$;
\item \textbf{case B} : squared exponential kernel with covariance function
$\rho(r)=\exp(-r^{2}/10^{2})$
\item \textbf{case C} : AR(4) process with poles $\exp(-0.02\pm0.05\cdot2\pi i)$,
$\exp(-0.10\pm0.30\cdot2\pi i)$, corresponding to spectral peaks
at $f=0.05$ and $f=0.30$.
\end{itemize}

Repeated irregular samples are drawn from this process as follows:
(i) a random draw of $N$ observations of the process is generated
from the prescribed covariance; (ii) samples are random selected selected
with probability $1/T$, where $T$ is the average sampling interval.
We use $N=1000$ and $T=5.$ Parameters are estimated for the AR($8$)
model for case A and B, and an AR($4$) model for case C. The motivation
for the lower order for case C is to create a best scenario case for
the ML estimator, by matching the estimated model order with the actual
model order. For the ML estimator, the ME increases more strongly
with increasing model order compared to PMEAN.

\begin{figure}[tbh]
\includegraphics[width=3.25in,height=1.99in]{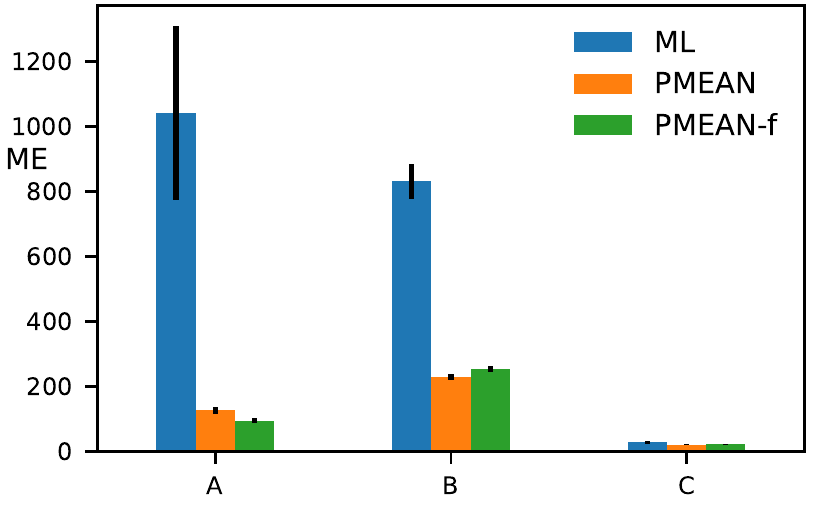}\protect\caption{\label{fig:ME-per-case}Model Error ME for the kernel case studies
for the ML and PMEAN estimators. The
posterior mean estimator PMEAN achieves a significant error reduction
for cases A and B. The ME is the average over 400 simulations.}
\end{figure}

The average model error over $S=400$ simulation runs is given in
figure \ref{fig:ME-per-case}. We draw the following main conclusions
from this experiment: (i) The Bayesian PMEAN estimator significantly
reduces the model error compared to ML for case A and B. The ME is
reduced by a factor of 8.2 for case A, by a factor of 3.6 for case
B; (ii) All estimators perform well for case C, which shows that PMEAN
is successful at suppression of spurious peaks in cases A and B, while
at the same time it correctly estimates a peak when it occurs in the
true spectrum.

In addition to PMEAN, we also report the results for the posterior
mean with a flat prior on the partial autocorrelations $\phi_{i}$
(PMEAN-f). The results for PMEAN-f are comparable to those of PMEAN,
showing that results are not sensitive to the kind of uninformative
prior used. Since this result holds equally for other reported simulation
experiments, PMEAN-f is not reported separately elsewhere.

\begin{figure}[tbhp]
\includegraphics{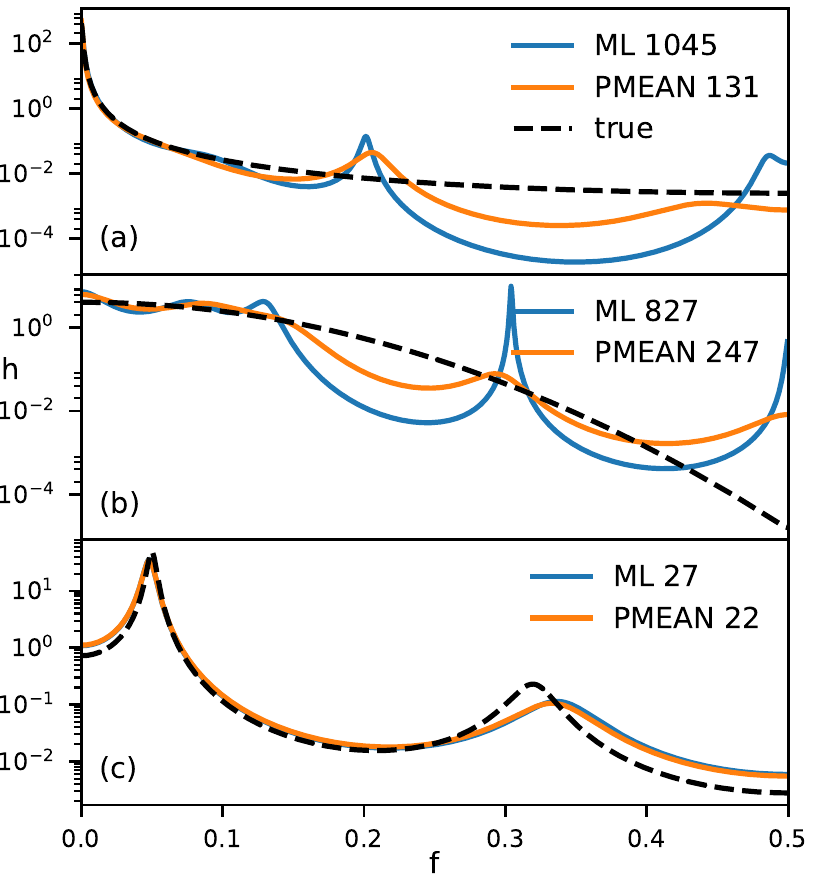}\protect\caption{\label{fig:cases-est-spectra}Estimated power spectra for representative
simulation runs for the kernel case studies, compared to the true
kernel spectrum. The legend shows the model error ME for each estimate.}
\end{figure}

Spectral estimates for representative simulation draws are given in
figure \ref{fig:cases-est-spectra} in comparison to the true spectrum
along with the resulting model error ME. For case A, the spectral estimate
at higher frequencies for ML estimate has an error of close to 3 orders
of magnitude, resulting in a large ME of 1045, while PMEAN is much
more accurate in this frequency range. Also for case B, we observe that
PMEAN is successful at suppression of spurious peaks compared to ML.
Finally, both estimators accurately model the true spectral peak when
it occurs in the true spectrum in case C.

\begin{figure}[tbh]
\includegraphics{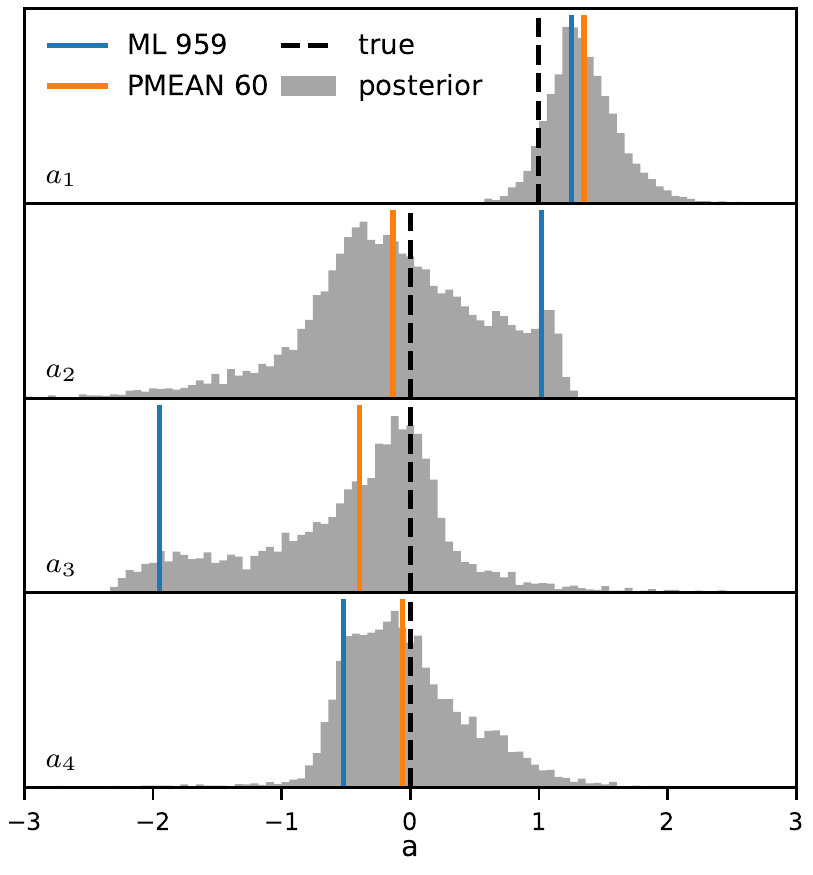}\protect\caption{\label{fig:post-case-A}Posterior distribution and point estimates
for AR parameters $a_{1}$ to $a_{4}$ (top to bottom) for an estimated AR(8) model
for case A. The legend shows the model error ME for each estimate.}
\end{figure}

Figure \ref{fig:post-case-A} shows the posterior distribution and
point estimates for AR parameters $a_{1}$ through $a_{4}$ for an
estimated AR(8) model for case A. Higher order coefficients are omitted
because they show a similar pattern. The posterior for the (non-zero)
parameter $a_{1}$ is quite narrow. Both the ML and PMEAN estimates
are accurate for this parameter. However, we observe a wide posterior
for the higher order coefficients, roughly centered around the true
value of $0$. For these coefficients, the posterior mean, being the
center of mass of the distribution, more robustly estimates a value
closer to $0$ because it takes into account the entire distribution,
thus resulting in a more accurate model. Conversely, the maximum likelihood
estimator does not take into asccount the entire shape of the distribution,
but only the maximum, and thus is more prone to shift
dramatically due to small random variations, resulting in a larger
error.

For AR(8) model estimation in cases A and B, the average computation
time for the ML estimator is $2.3$ seconds, while PMEAN takes $230$
seconds. Hence, ML is the best option kernel learning when its speed is
required. However, the computation time for PMEAN is reasonable
and its significantly greater accuracy makes it the algorithm of choice
in case the computational resources are available. Furthermore, our
results can motivate future work by accelerating the sampling, e.g.
by a specialized derivative computation instead of relying on the
Stan autodiff algorithm; usage of the DIAG-PRE-KAL algorithm;
and faster sampling algorithms such as variational inference.

\subsection{\label{sub:Performance-range}Performance under range of process
parameters}

\begin{figure}[tbh]
\includegraphics[bb=10bp 12bp 205bp 205bp,clip,scale=0.6]{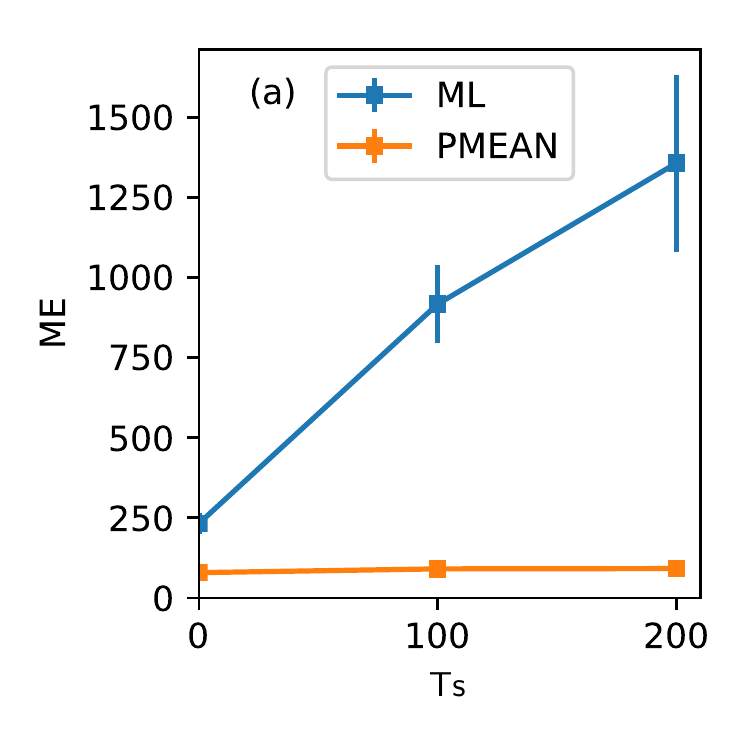}\includegraphics[bb=25bp 12bp 205bp 205bp,clip,scale=0.6]{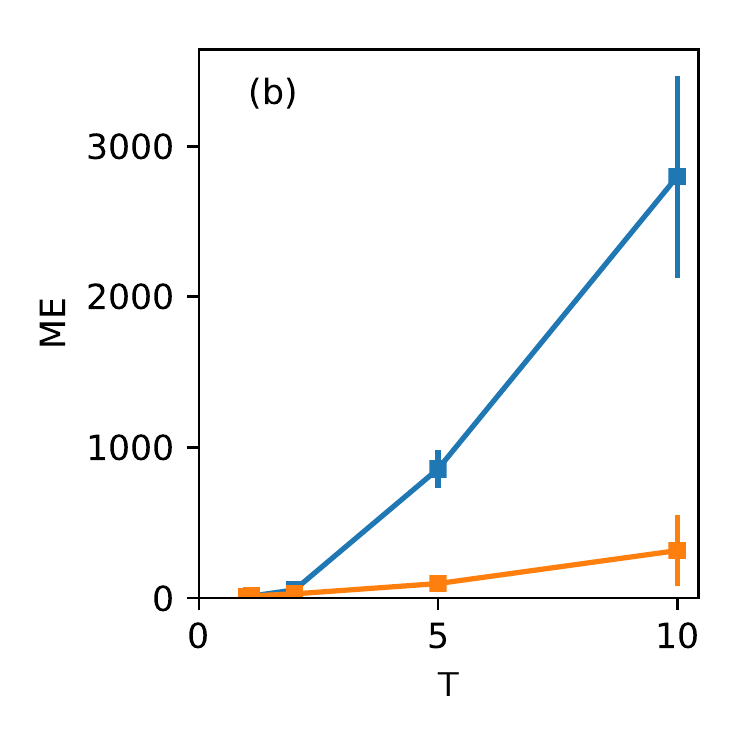}
\includegraphics[bb=10bp 12bp 205bp 205bp,clip,scale=0.6]{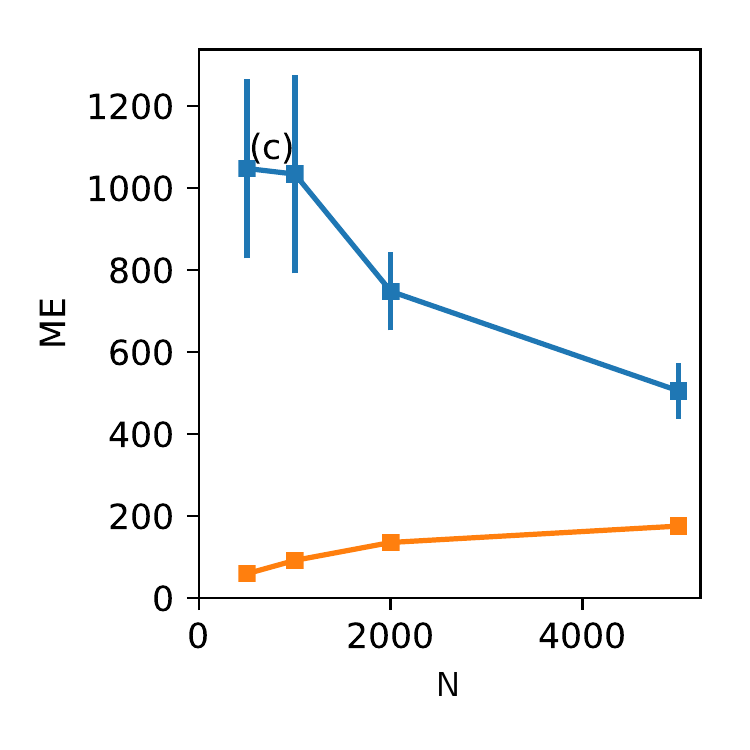}\includegraphics[bb=25bp 12bp 205bp 205bp,clip,scale=0.6]{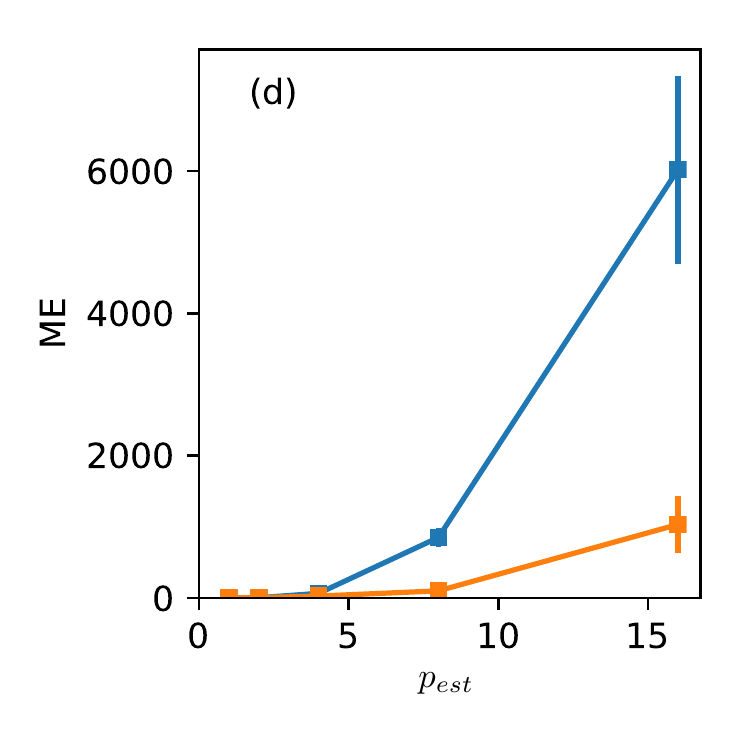}
\protect\caption{\label{fig:me-sweep}Model error ME of ML and PMEAN estimators as
a function of correlation length $T_{s}$ \textbf{(a)}; average sampling
time $T$ \textbf{(b)}; measurement time $N$ \textbf{(c)}; and model
order $p_{est}$\textbf{(d)}.}
\end{figure}

We report the results of a simulation experiment for a range of process
parameters, using case A from section
\ref{sub:Case-study} as a reference. When varying the sampling interval
$T_{s}$, the measurement time is increased in proportion to $T_{s}$
so that the average number of available samples $N_{a}$ remains constant. The results are given in figure \ref{fig:me-sweep}

Figure \ref{fig:me-sweep}a) shows that the ML estimator has a larger model error with increasing
correlation length $T_{s}$, corresponding to a larger dynamic range
in the frequency domain; increasing average sampling time $T$ (\ref{fig:me-sweep}b); and
increasing estimated model order $p_{est}$ (\ref{fig:me-sweep}c).

Conversely, the model
error decreases with increasing measurement time $N_a$ (\ref{fig:me-sweep}d), indicating
that the estimator converges to an accurate result as the number of
observations increases. Note that the model error is scaled with $N_a$.
Hence, the unscaled error decreases faster than $1/N_a$. This is caused
by finite sample effects, which are not described by asymptotic theory.
In the asymptotic regime, the model error is independent of $N_a$.

\section{\label{sec:monsoon}Monsoon rainfall variability data}

We investigate long-term monsoon rainfall variability based on radiometric-dated,
speleothem oxygen isotope $\delta^{18}O$ data \cite{sinha2015trends}.
This data enables evaluation of climate variability on a much larger
time scale than the few decades of precipitation recorded by meteorologists.
The data is intrinsically irregularly sampled, because it is formed
by natural deposition rather than experimenter controlled sampling.
For the same reason, irregular sampling occurs for many other long-term
climate records as well, e.g. ice core data \cite{petit1999climate}.

Speleothem (cave formation) data as studied across various locations can have a range
of average sampling rates. The current dataset is particularly suitable
for algorithm benchmarking because it has a higher average sampling
rate than datasets collected from other locations. This allows us
to study algorithm performance as a function of sampling rate by subsampling
the original data, and comparing the results to estimates obtained
from the full dataset.

The oxygen isotope data consists of $N=1848$ irregularly sampled
observations of $\delta^{18}O$ anomalies over a time span of $2147$
years, resulting in average sampling interval of $T_{0}=1.16$ years.
We convert the irregularly sampled data to a regularly sampled grid
with missing data with a grid spacing of $T_{g}=2$ years and use
this data to estimate AR(8) models. Because of the high sampling rate
and selected grid spacing, on 13\% of samples on the regular grid
are missing when all samples are used, and we can consequently estimate a reliable reference
model. This is confirmed by the fact that both the ML and PMEAN are
practically identical, with $\mbox{ME}<1$ between the two estimates.

We proceed to increase the average sampling interval $T$ by random
subsampling to create lower-fidelity datasets as they may be observed
at other locations. The sampling interval $T$ is expressed in grid
time steps, so that $T=1$ corresponds to no missing data, as in the
simulation experiments.

\begin{figure}[tbh]

\includegraphics{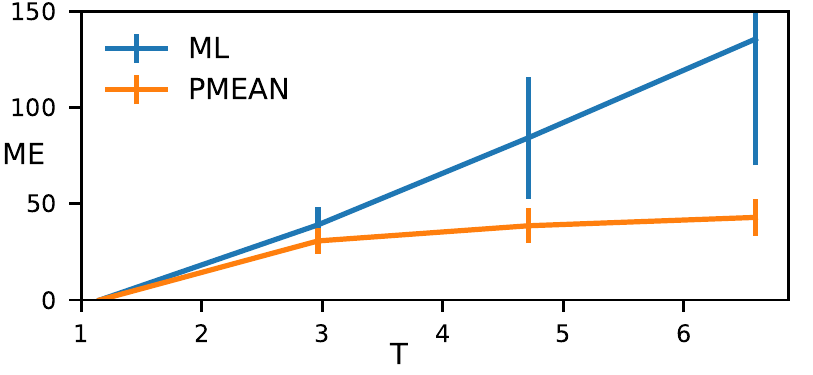}
\includegraphics{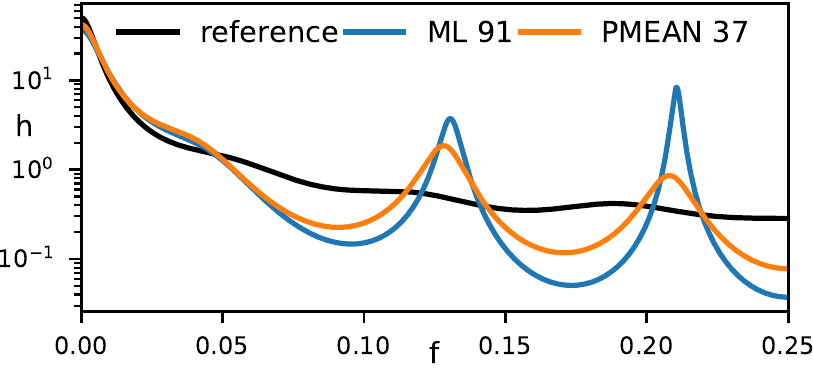}
\protect\caption{\label{fig:monsoon} Estimator performance
on randomly subsampled speleothem $\delta^{18}O$ data. As shown by the 
model error ME as a function of average sampling interval $T$ (a), the posterior mean
PMEAN is substantially more accurate than ML with increasing $T$. As in the simulation experiments,
the ML estimate shows spurious peaks which are strongly reduced in
the PMEAN estimate (b), resulting in a lower ME (37 vs 91)}
\end{figure}

For each subsampling rate, we generate $20$ different signals by
repeated random subsampling. The error of estimated AR($8$) models
compared to the reference model as a function of $T$, as well as estimated power spectra for $T=6.6$ are given in figure \ref{fig:monsoon}.

Consistent with
the simulation results, we observe that both estimators are accurate
for low $T$, but that the PMEAN estimator has a substantially lower
error as $T$ increases. Also, PMEAN successfully suppresses the spurious
peaks at higher frequencies.

\section{Concluding remarks}

The reported algorithms and results are directly relevant to kernel
learning when using Linear Gaussian Markov models for scalar index
variables or time series. Furthermore, the results can guide a wide
range of related research, as we motivate below.

First, we report that the posterior
mean based on the reference prior yields considerably
more accurate models than the ML estimate,
which strongly deteriorates for complex models.
It is expected that this result will generalize to other kernels.
Considering the trend towards more complex models, this is
a key result for the field of kernel learning.

Second, the Linear
Gaussian Markov model can be used to model data with multidimensional
index variables by using it to compose a separable multidimensional covariance function
\cite{gilboa2015scaling}.

\bibliography{references}
\bibliographystyle{icml2018}

\end{document}